\documentclass[10pt,twocolumn,letterpaper]{article}

\usepackage[pagenumbers]{memetag_wacv}
\usepackage{graphicx}
\usepackage{amsmath}
\usepackage{amssymb}
\usepackage{booktabs}
\usepackage{array}
\usepackage[round, sort, numbers]{natbib}
\setcitestyle{square}

\usepackage[breaklinks,hidelinks]{hyperref}

\def\confName{WACV}
\def\confYear{2026}

\begin{document}

\title{MemeTAG: Keyword-Driven Meme Classification through Tag Embedding Reconstruction}
\author{
Akshit Sharma \qquad Prashant W Patil\\
Indian Institute of Technology Guwahati\\
{\tt akshitsharma.rs@gmail.com, pwpatil@iitg.ac.in}
}

\maketitle

\begin{abstract}
 The proliferation of harmful internet memes poses a significant societal threat, yet their automated classification remains a formidable algorithmic challenge due to the nuanced, multimodal nature of their content. To address this, we introduce MemeTAG, a novel dual-objective framework that pioneers a keyword-aware approach to meme classification. Our core innovation is a two-part semantic guidance mechanism: first, we leverage a pretrained Vision-Language Model to generate a set of descriptive keywords, that capture the high-level semantics. Second, we introduce the Aggregated Tag Inference Network (ATIN), an attention-based module that distills these keywords into a single, rich semantic embedding. This embedding serves as a target for a novel auxiliary reconstruction loss, which compels the model to learn deeply aligned visual and textual features. This approach, combined with an efficient three-stage training strategy, establishes a new state-of-the-art on the HarMeme, Hateful Memes Challenge (HMC), and PrideMM datasets, decisively outperforming existing state-of-the-art methods.
\end{abstract}

\section{Introduction}
Memes---multimodal online messages combining images and brief textual elements---have become ubiquitous in digital communication, enabling the rapid dissemination of humor, cultural commentary, societal viewpoints, and shared emotional expressions. Their ease of creation, widespread accessibility, and inherent virality allow them to swiftly shape narratives, influence public discourse, and reflect collective societal attitudes. However, this powerful communication medium is increasingly exploited for disseminating harmful narratives, including hate speech, discriminatory ideologies, and misinformation, posing significant risks to online safety and social cohesion~\cite{castano2021internet, davidson2017automated}. Accurately classifying memes within this multimodal context presents unique technical challenges. The interplay between visual and textual components can be subtle, context-dependent, and culturally nuanced. Individually innocuous or neutral images and text can combine to become explicitly harmful or offensive, requiring sophisticated multimodal models capable of understanding the underlying semantic interactions~\cite{perifanos2021multimodal, sharma2023memex}. Current automated approaches, despite increasing sophistication, often struggle to capture such nuanced multimodal relationships due to their reliance on generalized representations or insufficient semantic alignment between modalities. Thus, improving meme classification demands a deeper integration of visual and textual modalities, informed by contextual semantic signals.
\begin{figure*}[t]
\centering \includegraphics[width=0.97\textwidth]{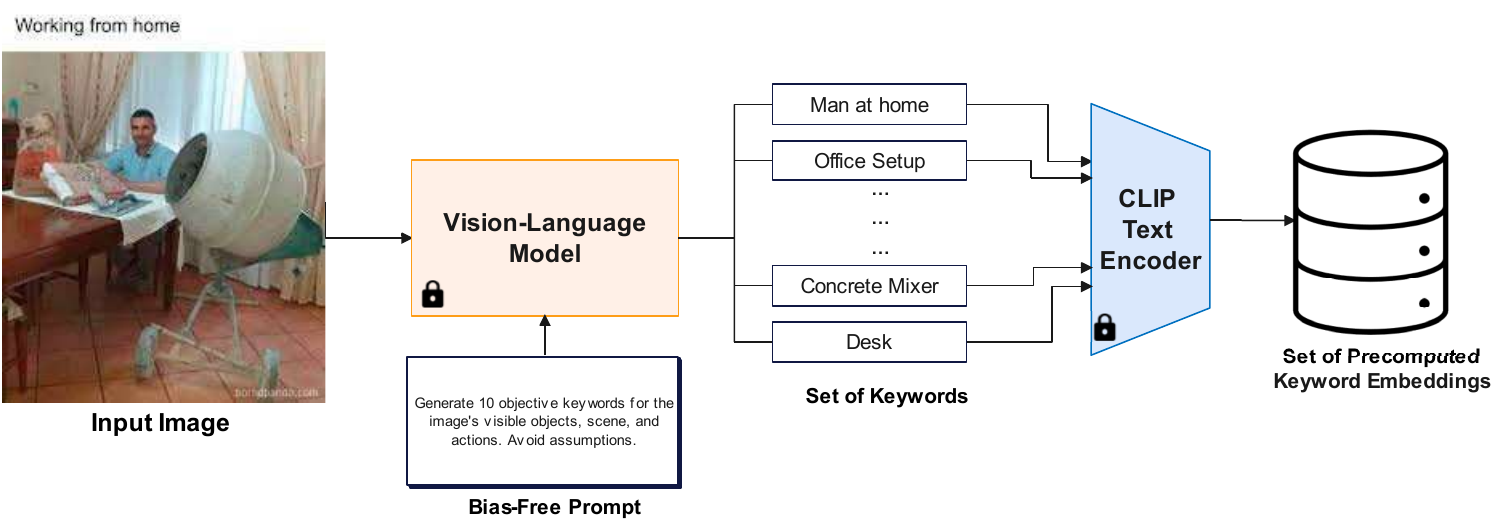}
\vspace{-2mm}
    \caption{Process of generating a set of keyword embeddings via frozen Vision-Language Model (VLM) using a neutral-language prompt. These generated keyword embeddings are used to train the Aggregated Tag Inference Network (ATIN)}
    \label{fig:keyword-embedding-generation}
    \vspace{-2mm}
\end{figure*}

In this work, we introduce MemeTAG, a dual-objective multimodal framework designed for an in-depth understanding of Internet memes. MemeTAG promotes deeper alignment between visual and textual features using semantic keywords derived from the image, enhancing overall semantic coherence and improving classification performance. Additionally, we propose an efficient three-stage training strategy that significantly reduces the computational overhead typically associated with training large-scale models, without compromising their representational power. We evaluate MemeTAG extensively on three challenging benchmark datasets: PrideMM~\cite{shah2024memeclip}, focused on LGBTQ+ community memes, and HarMeme~\cite{pramanick2021detectingharmfulmemestargets}, containing COVID-19-related hateful content and Hateful Memes Challenge (HMC)\cite{kiela2020hateful}. Empirical results confirm that MemeTAG outperforms state-of-the-art methods on key metrics including accuracy, AUROC, and F1 score, showcasing its effectiveness and efficiency in addressing complex multimodal meme classification tasks. Our primary contributions are summarized as follows:
\begin{itemize}
\item We propose MemeTAG, a dual-objective multimodal framework enabling enhanced semantic alignment and thereby accurate meme classification through keyword-guided visual and textual integration.
\vspace{1mm}
\item As part of MemeTAG, We introduce a novel auxiliary reconstruction task  to guide the model by enforcing semantic alignment between modalities, as shown in the ablation study Sec.~\ref{sec:atin-ablation}
\vspace{1mm}
\item We perform extensive evaluations on the challenging PrideMM, HarMeme, and HMC datasets, benchmarking MemeTAG against a broad suite of multimodal baselines---and demonstrate that it consistently achieves state-of-the-art performance on complex meme classification tasks.
\end{itemize}

\begin{figure*}[t]
    \centering \includegraphics[width=0.9\textwidth]{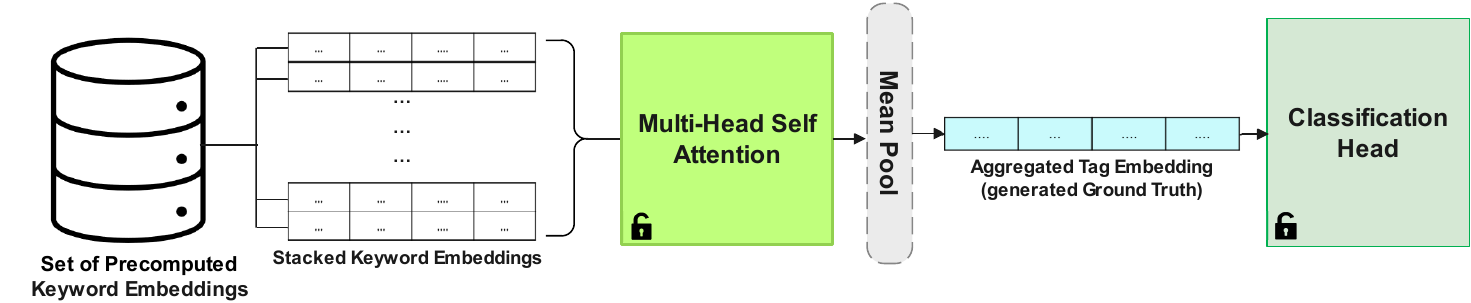}
    \vspace{-2mm}
    \caption{Architecture of the proposed Aggregated Tag Inference Network (ATIN). The aggregated tag embedding from the trained ATIN serves as the ground in the auxiliary reconstruction task of MemeTAG}
    \label{fig:atin-architecture} \vspace{-2mm}
\end{figure*}
\section{Literature Review}

The rise of harmful content on social media has spurred research into classifying internet memes---which blend humor with potentially offensive messaging \cite{Hee2024a, Uyheng2020}. Early multimodal classifiers combined image and text features to detect harmful memes \cite{Suryawanshi2020}, while Xu et al.~\cite{Xu2022} exposed challenges in sentiment analysis of metaphorical memes. To handle multilingual and code-mixed memes, weighted ensembles~\cite{Hossain2022} and the CM-Off-Meme multitask model~\cite{Kumari2024} were proposed. Explainability was advanced by HateXplain's BERT-based framework~\cite{Mathew2021}, and Sudarshan et al.~\cite{Sudarshan2022} utilized VisualBERT and RoBERTa for targeted hate detection. Recent work has increasingly shifted towards leveraging the powerful representations from large pre-trained vision-language models. An early example, MOMENTA, was among the first to incorporate CLIP features, employing a two-perspective framework that analyzes memes at both global and local (e.g., entity-level) scales to detect harm and identify its target \cite{momenta}. The success of such models highlighted the challenge of effectively adapting large models to the specific, often data-scarce, domain of meme analysis. To address this, CLIP-Adapter proposed adding a single, lightweight adapter layer to CLIP's frozen features, which proved effective in leveraging CLIP's prior knowledge while minimizing overfitting \cite{radford2021clip}. Building on this, MemeCLIP further refined the adaptation process by using separate, modality-specific adapters for image and text features, coupled with a cosine similarity-based classifier to mitigate class imbalance issues prevalent in meme datasets \cite{shah2024memeclip}. Beyond adaptation, significant effort has focused on developing more sophisticated multimodal fusion mechanisms. HateCLIPper explicitly models cross-modal interactions by computing a feature interaction matrix from CLIP embeddings, capturing pairwise correlations between visual and textual feature dimensions to achieve state-of-the-art results on the Hateful Memes Challenge \cite{Kumar2022}. MemeFier proposed a dual-stage fusion architecture for meme understanding, aligning CLIP token/patch features followed by inter-model dependencies using transformer and caption-supervision regularizer besides external demographic cues. \cite{10.1145/3591106.3592254} In a distinct approach, ISSUES utilizes textual inversion to map an entire meme image to a learned pseudo-word within CLIP's textual embedding space, thereby unifying the multimodal problem within the language domain to better capture subtle semantics \cite{issues}. Concurrently, efforts like GuardHarMem have pushed for more nuanced evaluation by introducing datasets with fine-grained harm labels (e.g., mockery, racism) and strong multimodal baselines that fuse image, OCR, and caption data \cite{el2025guardharmem}. Other recent work integrates large language models (LLMs) more directly for reasoning. Sharma et al.~\cite{Sharma2023} introduce the VECTOR framework for contextual entity roles, and Cao et al.~\cite{Cao2023} present PromptHate with illustrative examples. Expanding on these reasoning-based approaches, M3H injects commonsense and figurative knowledge by using an LLM to generate an enriched explanation of a meme's implicit meaning, which is then used for classification \cite{m3h}. The concept of prompting was further advanced by the Prompt-Enhanced Network (PEN), which learns internal, feature-level prompts to guide the model's perception, moving beyond static input prompts to a more dynamic, optimizable framework \cite{Liu2024}. These collective efforts showcase the field's rapid progression toward building robust, context-aware, and interpretable meme classification models.
\subsection{Vision-Language Models}
Vision-language models (VLMs) have rapidly advanced with the adoption of Transformers. Early attention-based systems like ``Show, attend and tell''~\citep{xu2015show} gave way to the Vision Transformer (ViT)~\citep{dosovitskiy2021imageworth16x16words} and CLIP's contrastive pre-training on hundreds of millions of image--text pairs (often from LAION)~\citep{radford2021clip, schuhmann2022laion5b, schuhmann2021laion400m}, enabling strong zero-shot transfer. Building on these, Flamingo~\citep{alayrac2022flamingo} pioneered few-shot learning, while the BLIP series~\citep{li2022blip, li2023blip2} and LLaVA~\citep{liu2023visual} advanced encoder--decoder and instruction-tuned designs. More recently, decoder-only Multimodal LLMs such as GPT-4V~\citep{openai2023gpt4v} and Gemini~\citep{team2023gemini}, along with open-source variants like Qwen2-VL~\citep{qwen2024qwen2}, and LLaMA 3.2-vision~\citep{meta2024llama32}, have scaled capabilities. Current efforts focus on human-aligned outputs via RL techniques (RLOO, DPO, PPO) in models such as MM-Eureka~\citep{meng2025mmeurekaexploringvisualaha}, MM-RLHF~\citep{zhang2025mm} and LMM-R1~\citep{peng2025lmmr1}.
\section{Proposed Methodology}
\label{sec:methodology}
The proposed MemeTAG model starts by using a VLM to generate semantic keywords from the meme's image, which are then embedded using CLIP's text encoder. These keyword embeddings flow into the ATIN module, an attention network that produces a single aggregated tag embedding ($Z_{\text{tag}}$) capturing salient visual semantics. In parallel, the main architecture extracts image and text features using CLIP, processes and fuses them into a multimodal representation ($f$, further refined to $f'$). This final representation ($f'$) is used for classification via an ArcFace-inspired head, while the intermediate fused representation ($f$) is guided by an auxiliary loss to reconstruct the semantic tag embedding ($z_{\text{tag}}$), ensuring alignment during training with a combined objective function.
\subsection{Keyword Inference and Embedding Generation}
Given an Internet meme that includes both an image and text, the first stage of our proposed MemeTAG framework is to \textbf{infer} a concise set of descriptive keywords that capture the image's semantic content. We utilize a Vision-Language Model (VLM) to generate these keywords from the image. To ensure uniformity across samples, we restrict each meme to exactly $10$ keywords, automatically selecting the first $10$ keywords generated by the pretrained VLM. In cases where fewer than $10$ keywords are produced, we pad the keyword set with a special token \textit{null}. Subsequently, embeddings for these keywords are computed using the frozen CLIP~\citep{radford2021clip} architecture, pre-computing and storing them across the entire dataset for efficiency. Let the set of keywords be denoted by
\begin{equation}
  K = \{ k_1, k_2, \dots, k_{10} \}
\end{equation}
and the corresponding embeddings by
\begin{equation}
 E = \{ \mathbf{e}_1, \mathbf{e}_2, \dots, \mathbf{e}_{10} \}
\end{equation}

\begin{figure*}[t!]
    \centering
    \includegraphics[width=0.99\textwidth, height=0.50\textwidth]{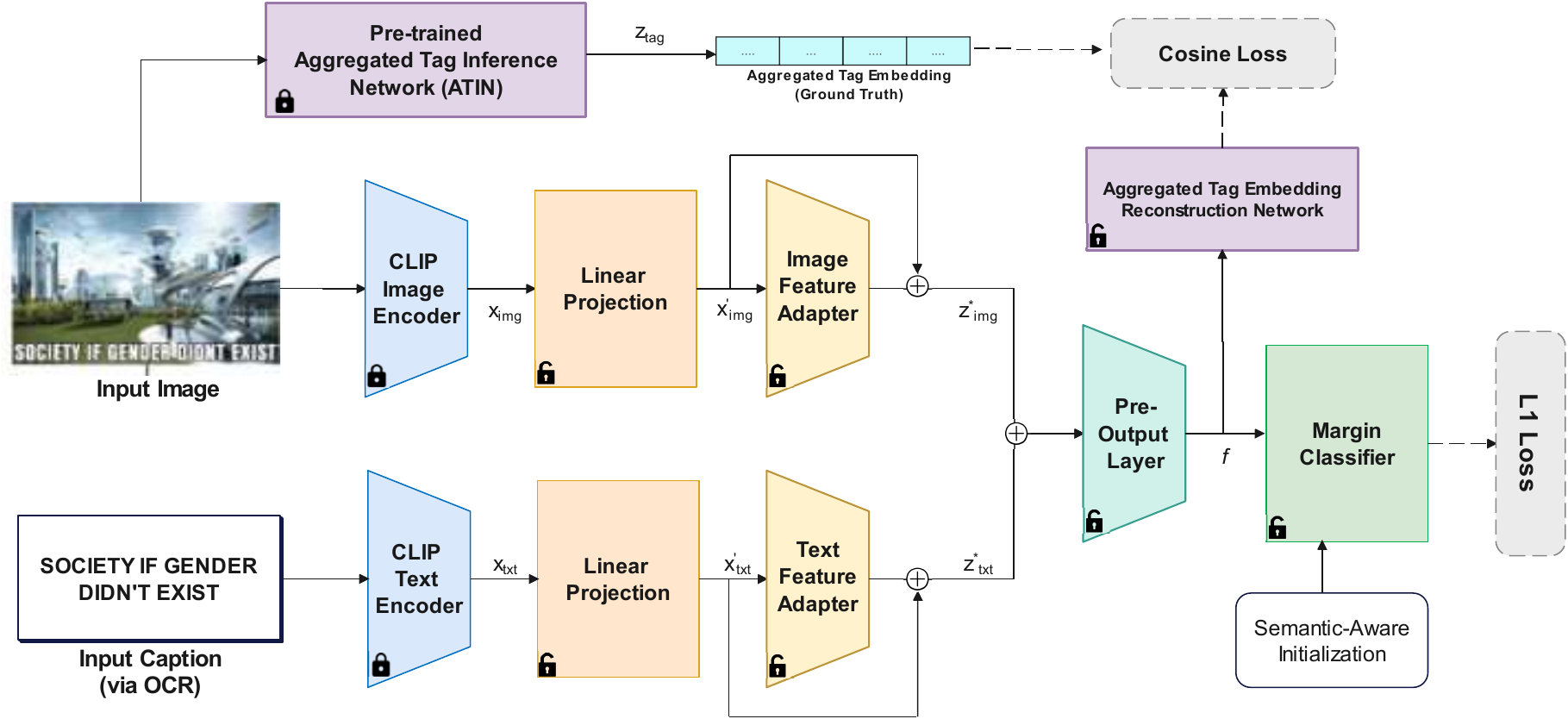}
    \vspace{-2mm}
    \caption{Architecture of the proposed MemeTAG model for keyword-aware meme classification. We make use of the trained Aggregated Tag Inference Network (ATIN) as shown in Figure \ref{fig:atin-architecture}}
    \label{fig:memetag-architecture}
\end{figure*}
\subsection{Aggregated Tag Embedding Construction via Attention Network}

The Aggregated Tag Inference Network (ATIN) is an attention-based module designed to distill a set of keyword embeddings, $E=\{e_{1},...,e_{10}\}$, into a single, rich semantic vector, $z_{tag}$. The architecture, shown in Figure \ref{fig:atin-architecture}, is centered around a Multi-Head Self-Attention (MHSA) mechanism. Unlike typical transformer-based aggregators, ATIN uses a learnable query vector ($q$) to probe the keyword embedding set ($E$). This allows the model to learn the most effective way to summarize the keywords. First, a context-aware representation, $E'$, is computed by applying the MHSA module, followed by a residual connection and layer normalization:
$$
E' = \text{LayerNorm}(E + \text{MHSA}(q, E, E))
$$
The final aggregated tag embedding, $z_{tag}$, is then produced by mean-pooling the resulting sequence of context-aware embeddings:
$$
z_{tag} = \text{MeanPool}(E') = \frac{1}{10} \sum_{i=1}^{10} e'_{i}
$$
As part of our three-stage training strategy, the ATIN module is trained independently by attaching a classification head to its output during the tag aggregation stage. Afterwards, its weights are frozen, and it is used to generate the ground-truth $z_{tag}$ embedding for each sample. This vector subsequently serves as the target for MemeTAG's auxiliary reconstruction objective, enforcing semantic alignment during the main model's training.
\subsection{MemeTAG Multimodal Architecture}
We propose MemeTAG, a multimodal neural architecture explicitly designed for keyword-aware meme classification, leveraging robust semantic embeddings from pre-trained CLIP encoders. Given an Internet meme comprising an image \( I \) and textual content \( T \), MemeTAG begins by extracting modality-specific embeddings: $\mathbf{x}_{\text{img}}\in\mathbb{R}^{d_{\text{img}}}$ from the frozen CLIP visual encoder and $\mathbf{x}_{\text{txt}}\in\mathbb{R}^{d_{\text{txt}}}$ from the frozen CLIP textual encoder.

To effectively integrate these modality-specific embeddings, MemeTAG first projects each embedding into a shared latent embedding space \(\mathbb{R}^{d_{\text{shared}}}\) via dedicated linear projection networks:
\begin{align} \mathbf{x}'_{\text{img}} &= \text{LinearProj}_{\text{img}}(\mathbf{x}_{\text{img}}) \\\,\,\,\,\,\,\,\,
    \mathbf{x}'_{\text{txt}} &= \text{LinearProj}_{\text{txt}}(\mathbf{x}_{\text{txt}})
\end{align} where \(\text{LinearProj}_{\text{img}}\) and \(\text{LinearProj}_{\text{txt}}\) are multi-layer linear transformations with dropout regularization. Following the projections, each embedding undergoes modality-specific adapter modules \(f_{\text{img}}\) and \(f_{\text{txt}}\), refining these embeddings for multimodal fusion:
\begin{equation}
\begin{aligned}
\mathbf{z}_{\text{img}} &= f_{\text{img}}\left(\mathbf{x}'_{\text{img}}\right); \,\,\,\,\,\,\,\,\,\,
\mathbf{z}_{\text{txt}} &= f_{\text{txt}}\left(\mathbf{x}'_{\text{txt}}\right).
\end{aligned}
\end{equation}

Subsequently, each adapted embedding is blended with its original projected counterpart using hyperparameter-defined ratios \(\alpha_{\text{img}}, \alpha_{\text{txt}}\in[0,1]\):
\begin{equation}
   \mathbf{z}^*_{\text{img}} = \alpha_{\text{img}}\mathbf{z}_{\text{img}} + (1 - \alpha_{\text{img}})\mathbf{x}'_{\text{img}}
\end{equation}
\begin{equation}
    \mathbf{z}^*_{\text{txt}} = \alpha_{\text{txt}}\mathbf{z}_{\text{txt}} + (1 - \alpha_{\text{txt}})\mathbf{x}'_{\text{txt}}
\end{equation}
These blended embeddings are individually normalized (using Layer Normalization) and subsequently fused via element-wise multiplication ($\odot$) to capture detailed cross-modal interactions, resulting in:
\begin{equation}
   \mathbf{f} = \operatorname{Norm}(\mathbf{z}^*_{\text{img}})\odot\operatorname{Norm}(\mathbf{z}^*_{\text{txt}})
\end{equation}
To further enhance representational capacity and introduce non-linearity, MemeTAG applies transformations through pre-output layers, composed of linear transformations, Rectified Linear Units (ReLU), and dropout, yielding:
\begin{equation}
   \mathbf{f}' = \text{Dropout}\left(\text{ReLU}\left(\mathbf{W}_{p}\mathbf{f} + \mathbf{b}_{p}\right)\right)
\end{equation}
where $\mathbf{W}_{p}\in\mathbb{R}^{d_{\text{shared}}\times d_{\text{shared}}}$ and $\mathbf{b}_{p}\in\mathbb{R}^{d_{\text{shared}}}$ are trainable parameters.
For classification, MemeTAG employs a margin-based classifier inspired by ArcFace. This computes class logits \(\ell\) by comparing the final representation $\mathbf{f}'$ with class weights $\mathbf{W}_c$ and incorporating an angular margin \( m \):
\begin{equation}
    \cos(\theta_j) = \frac{\mathbf{W}_{c,j}^\top\mathbf{f}'}{\|\mathbf{W}_{c,j}\|\|\mathbf{f}'\|+\epsilon} \end{equation}
\begin{equation}
    \ell_j = s \cdot \phi_j(\mathbf{f}') = \begin{cases} s \cdot \cos(\theta_j + m) & \text{if } j = y \\ s \cdot \cos(\theta_j) & \text{if } j \neq y \end{cases} \end{equation}
where $\mathbf{W}_c\in\mathbb{R}^{d_{\text{shared}}\times C}$ represents the classifier weight matrix ($\mathbf{W}_{c,j}$ being the weights for class $j$), \(y\) is the ground-truth class index, \(C\) is the total number of meme classes, \(s\) denotes a scaling factor, and \(\epsilon\) is a small constant ensuring numerical stability. Angular margin explicitly promotes discriminative embeddings,thus improving classification performance.

\subsection{Loss Functions}
The MemeTAG framework is optimized using two complementary loss functions:\\
\textbf{Classification Loss:} The main objective is to accurately classify Internet memes. This is achieved using a classification loss applied to the final representation $\mathbf{f}'$ via the margin-based logits $\ell_j$ derived from ArcFace. Formally, the classification loss $(\mathcal{L}_{\text{cls}})$ is the standard cross-entropy loss calculated using these logits:
\begin{equation}
  \mathcal{L}_{\text{cls}} = -\log\left(\frac{e^{\ell_y}}{\sum_{j=1}^{C} e^{\ell_j}}\right)
\end{equation}
where $y$ denotes the ground-truth class index.\\
\textbf{Auxiliary Reconstruction Loss:} To enforce semantic alignment between the multimodal features and the aggregated keyword information, an auxiliary objective is incorporated. This task aims to reconstruct the aggregated semantic tag embedding $\mathbf{z}_{\text{tag}}$ (produced by ATIN) from the fused multimodal feature $\mathbf{f}$ (obtained before the final pre-output layers). A reconstruction network (e.g., a simple MLP) takes $\mathbf{f}$ as input and outputs a reconstructed embedding $\hat{\mathbf{z}}_{\text{tag}}$. The reconstruction loss $\mathcal{L}_{\text{aux}}$ is then computed based on the cosine distance (1 - cosine similarity) between the target and the prediction:
\begin{equation}
    \mathcal{L}_{\text{aux}} = 1 - \frac{\mathbf{z}_{\text{tag}} \cdot \hat{\mathbf{z}}_{\text{tag}}}{\|\mathbf{z}_{\text{tag}}\| \, \|\hat{\mathbf{z}}_{\text{tag}}\|}
\end{equation}
where $\hat{\mathbf{z}}_{\text{tag}}$ is the reconstructed tag embedding produced by the reconstruction network. The overall objective function is a weighted sum of the classification and auxiliary losses:
\begin{equation}
    \mathcal{L} = (1-\lambda)\mathcal{L}_{\text{cls}} + \lambda \, \mathcal{L}_{\text{aux}}
\end{equation}
with $\lambda$ being a hyperparameter that balances the contribution of the two objectives.
\section{Experimental Details and Result Analysis}
In this section, we detail the dataset configurations for three datasets- PrideMM, HarMeme, and HMC, along with the training strategy and analysis of the results. Due to space constraints, additional experimental results and model implementation details are provided in the Supplementary Material.
\subsection{Dataset Configuration}
We evaluate MemeTAG on three multimodal meme benchmark: PrideMM, HarMeme, and HMC.  PrideMM \cite{shah2024memeclip} comprises 5,063 LGBTQ+--related memes annotated for four aspects---hate (no hate/hate), target (undirected/individual/community/organization), stance (neutral/support/oppose) and humor (no humor/humor)---with official train/test splits.  HarMeme (Harm-C) \cite{pramanick2021detectingharmfulmemestargets} contains 3,544 COVID-19--related memes for binary hateful/non-hateful classification, partitioned into 85\% train, 5\% validation and 10\% test.  The Facebook Hateful Memes Challenge dataset \cite{kiela2020hateful} comprises roughly 10,000 memes labeled hateful vs. non-hateful with official train, dev (seen/unseen) and test (seen/unseen) splits; we report our results on its test ``seen'' split.  In all three cases, MemeTAG is trained on the  train split and evaluated on the specified test split.

\subsection{Training Strategy}
We adopt a three-step training strategy designed to reduce computational overhead and enhance convergence stability:\\
\textbf{Precomputation Stage}: Let the dataset be denoted by
    \(\mathcal{D} = \{\bigl(I_n, T_n, \{\mathbf{k}_n^j\}_{j=1}^{10}\bigr)\}_{n=1}^N,\)
    where \(I_n\) is the \(n\)-th meme image, \(T_n\) the associated text, and \(\{\mathbf{k}_n^j\}_{j=1}^{10}\) represents the set of keyword embeddings corresponding to meme \(n\). For each sample \(n\) in the dataset, we first use the pretrained CLIP encoders to generate and store the image embedding \(\mathbf{x}_{\text{img}, n} \in \mathbb{R}^{d_{\text{img}}}\), the text embedding \(\mathbf{x}_{\text{txt}, n} \in \mathbb{R}^{d_{\text{txt}}}\), and the set of keyword embeddings \(\{\mathbf{k}_n^j \in \mathbb{R}^{d_{\text{kw}}}\}_{j=1}^{10}\). Storing these frozen embeddings serves as the foundation for subsequent network components, mitigating the cost of repeated on-the-fly inference.\\
\textbf{Tag Aggregation Stage}: Using the precomputed keyword embeddings \(\{\mathbf{k}_n^j\}_{j=1}^{10}\) for each sample \(n\), we train the Aggregated Tag Inference Network (ATIN), as detailed in Section~3.2. This module learns attention weights \(\{\alpha_j\}_{j=1}^{10}\) specific to each set of 10 keywords and generates a context-enriched tag embedding \(\mathbf{z}_{\text{tag}, n}\) for each sample via: \begin{equation}
        \mathbf{z}_{\text{tag}} \;=\; \sum_{j=1}^{10} \alpha_j\, \mathbf{k}_n^j
    \end{equation}
This process captures crucial semantic cues from the image keywords for each meme \(n\). \\
\textbf{MemeTAG Training Stage}: Finally, leveraging the precomputed aggregated tag embeddings \(\mathbf{z}_{\text{tag}, n}\) and the frozen CLIP embeddings \(\mathbf{x}_{\text{img}, n}\), \(\mathbf{x}_{\text{txt}, n}\) for each sample \(n\), we train the main MemeTAG architecture. This stage focuses on learning to jointly integrate the image, text, and tag signals effectively for the classification task.\\ By initializing with these precomputed representations, we minimize computational costs while ensuring the model learns robust, semantically aligned multimodal features. This hierarchical training strategy significantly reduces total compute time compared to typical end-to-end training, fosters stable convergence, and yields powerful multimodal embeddings optimized for accurate keyword-aware meme classification.

\subsection{Result Analysis}
Tables \ref{tab:pridemm-results}, \ref{tab:harmeme-results} and \ref{tab:hmc-results} summarize our experimental results on the PrideMM, HarMeme and HMC datasets, respectively. For the PrideMM dataset, we used the predefined train/validation/test splits (ratio 85/5/10), while for HarMeme and HMC we used its established predefined splits. We evaluated various unimodal and multimodal methods using Accuracy, AUC (Macro), and F1-Score (Macro) as evaluation metrics. For PrideMM, these metrics were computed across its four distinct tasks: Hate, Target, Stance, and Humor.

On the PrideMM dataset, our proposed model, MemeTAG, consistently achieved the best performance among the compared methods. Specifically, it obtained the highest Accuracy and F1 scores on the Hate, Target, and Humor tasks, along with the highest AUC and F1 score on the Stance task. Furthermore, analysis of the CLIP baselines on PrideMM revealed that the Image-Only variant consistently outperformed the Text-Only variant, highlighting the importance of visual features for this dataset. Multimodal methods that combine both image and text generally yielded further performance improvements over unimodal approaches.\\ On the HarMeme dataset, MemeTAG again emerged as the top-performing model, surpassing other methods evaluated, with HateCLIPper being the closest competitor in performance. A similar trend regarding modality importance was observed for the basic CLIP methods on HarMeme: the Image-Only variant's accuracy exceeded that of the Text-Only variant, and standard multimodal CLIP and CLIP-Adapter models provided additional gains. Other competitive multimodal frameworks, including GuardHarMem, MemeCLIP and ISSUES delivered strong results.\\
On the Hateful Memes Challenge (HMC), MemeTAG achieves an accuracy of 74.10\% and an AUROC of 82.17\%, as shown in Table~\ref{tab:hmc-results}. MemeTAG outperforms the second-best baseline, MemeCLIP (72.69\% Acc., 81.68\% AUROC), by 1.41\% Accuracy and achieving a higher AUROC (82.17\% vs.\ 81.68\%). Compared to ISSUES (70.80\% accuracy, 79.03\% AUROC), MemeTAG yields 3.30\% accuracy improvement.This demonstrates MemeTAG's superior alignment of visual and textual cues under HMC's adversarial setup.
\begin{table*}[ht]
\centering
\caption{Performance analysis of the proposed MemeTAG with existing state-of-the-art methods on PrideMM dataset}
\label{tab:pridemm-results}
\vspace{-2mm}
\resizebox{\textwidth}{!}{\begin{tabular}{|l|ccc|ccc|ccc|ccc|}
\hline
\multicolumn{1}{|c|}{\textbf{Method}} & \multicolumn{3}{c|}{\textbf{Hate}} & \multicolumn{3}{c|}{\textbf{Target}} & \multicolumn{3}{c|}{\textbf{Stance}} & \multicolumn{3}{c|}{\textbf{Humor}} \\ \cline{2-13} & \textbf{Acc.} & \textbf{AUC.} & \textbf{F1} & \textbf{Acc.} & \textbf{AUC.} & \textbf{F1} & \textbf{Acc.} & \textbf{AUC.} & \textbf{F1} & \textbf{Acc.} & \textbf{AUC.} & \textbf{F1} \\ \hline \hline CLIP Text-Only & 69.030 & 74.990 & 69.110 & 50.720 & 73.210 & 48.160 & 50.990 & 67.910 & 49.950 & 69.940 & 71.230 & 62.880 \\
CLIP Img-Only  & 70.550 & 80.710 & 73.820 & 61.290 & 81.180 & 58.630 & 61.440 & 77.810 & 58.330 & 76.230 & 82.880 & 73.040 \\
CLIP           & 72.960 & 80.770 & 72.880 & 61.480 & 82.090 & 58.890 & 59.720 & 77.310 & 58.270 & 77.180 & 80.820 & 73.910 \\
GuardHarMem  \cite{el2025guardharmem} & 70.020 & 75.740 & 67.660 & 65.480 & 75.050 & 49.250 & 52.660 & 71.400 & 52.010 & 71.600 & 73.520 & 79.790 \\
CLIP-Adapter \cite{radford2021clip}  & 73.210 & 81.140 & 73.090 & 61.870 & 82.290 & 58.470 & 59.830 & 77.590 & 58.390 & 77.510 & 81.240 & 73.990 \\
MOMENTA  \cite{momenta}      & 72.540 & 78.810 & 71.970 & 57.880 & 79.420 & 53.490 & 56.550 & 74.710 & 55.820 & 75.110 & 78.090 & 72.630 \\
HateCLIPper \cite{Kumar2022}   & 75.840 & 83.330 & 74.260 & 63.410 & 80.880 & 57.080 & 63.680 & 78.610 & 57.590 & 76.240 & 83.790 & 75.580 \\
ISSUES \cite{issues}         & 75.330 & 84.390 & 74.610 & 62.180 & 78.860 & 58.390 & 59.940 & 77.980 & 57.890 & 79.370 & 85.090 & 76.680 \\

MemeCLIP \cite{shah2024memeclip}      & 75.960 & 83.910 & 75.230 & 64.780 & 81.930 & 58.650 & 62.320 & 79.910 & 58.580 & 79.110 & \textbf{85.600} & 75.760 \\

\hline
\textbf{MemeTAG}        & \textbf{77.320} & \textbf{84.690} & \textbf{77.310} & \textbf{67.130} & \textbf{85.880} & \textbf{66.630} & \textbf{63.910} & \textbf{80.740} & \textbf{61.840} & \textbf{80.470} & 85.310 & \textbf{79.830} \\ \hline \end{tabular}} \end{table*}

\begin{table}[ht]
\centering
\caption{Performance analysis of the proposed MemeTAG with existing state-of-the-art methods on HarMeme dataset}
\label{tab:harmeme-results} \begin{tabular}{|l|ccc|}
\hline
\textbf{Method} & \textbf{Acc.} & \textbf{AUC.} & \textbf{F1} \\ \hline \hline CLIP Text-Only & 73.880          & 79.520          & 71.710          \\
ATIN (Baseline) & 74.240 & 84.150 & 73.260 \\ \hline
CLIP Img-Only  & 79.460          & 89.120          & 78.710          \\ \hline CLIP           & 81.770          & 87.590          & 80.740          \\
CLIP-Adapter \cite{radford2021clip}   & 82.490          & 87.810          & 81.130          \\
MOMENTA  \cite{momenta}        & 82.750          & 88.040          & 81.690          \\
HateCLIPper \cite{Kumar2022}    & 83.910          & 91.050          & 83.470          \\
M3H \cite{m3h} & 79.420 & 78.180 & 84.690 \\
ISSUES \cite{issues}         & 81.830          & 91.980          & 80.880          \\

MemeCLIP \cite{shah2024memeclip}       & 82.720          & 92.070 & 81.720          \\
GuardHarMem \cite{el2025guardharmem} & 84.970 & 90.170 & 79.690 \\ \hline
\textbf{MemeTAG}        & \textbf{85.540} & \textbf{92.230}          & \textbf{84.810} \\ \hline \end{tabular}
\end{table}

\begin{table}[ht]
    \centering
    \caption{Performance analysis of the proposed MemeTAG with existing state-of-the-art methods on HMC dataset}
    \label{tab:hmc-results}
    \begin{tabular}{|l|c|c|}
        \hline
        \textbf{Model} & \textbf{Acc. (\%)} & \textbf{AUROC (\%)} \\
        \hline
        Image-Grid (ResNet152) & 50.00 & 50.00 \\
        Image-Region (R-CNN)   & 57.00 & 52.00 \\
        \hline
        Text-BERT              & 51.10 & 53.39 \\
        ATIN (Baseline)        & 57.40 & 61.60 \\
        \hline
        Late Fusion            & 56.50 & 54.07 \\
        Concat BERT            & 52.10 & 53.94 \\
        MMBT-Grid              & 50.00 & 50.00 \\
        MMBT-Region            & 57.00 & 52.00 \\
        ViLBERT  \cite{lu2019vilbert}             & 55.30 & 52.29 \\
        VisualBERT \cite{li2019visualbertsimpleperformantbaseline}          & 56.40 & 54.05 \\
        ViLBERT CC             & 54.20 & 52.79 \\
        VisualBERT COCO        & 60.10 & 61.34 \\
        MOMENTA \cite{momenta} & 58.30 & 62.66 \\
        M3H \cite{m3h}         & 60.20 & 69.56 \\
        GuardHarMem \cite{el2025guardharmem} & 61.20 & 68.11 \\
        HATECLIPper \cite{Kumar2022} & 68.31 & 76.46 \\
        ISSUES  \cite{issues}              & 70.80 & 79.03 \\
        MemeCLIP \cite{shah2024memeclip} & 72.69 & 81.68 \\
        \hline
        \textbf{MemeTAG} & \textbf{74.10} & \textbf{82.17} \\
        \hline
    \end{tabular}
\end{table}

\begin{table*}[t]
    \centering
    \caption{Comparison of MemeTAG with several state-of-the-art LLMs for Harmful Meme classification on Harmeme dataset}
    \label{tab:llm-comparison}
    \setlength{\tabcolsep}{7pt}
    \vspace{-2mm}
    \begin{tabular}{@{} l *{5}{w{c}{4ex}} ccc @{}}
        \toprule
        \textbf{Model} & \multicolumn{2}{c}{\textbf{Supported Modalities}} & \multicolumn{3}{c}{\textbf{Input Modalities Used}} & \textbf{Accuracy (\%)} & \textbf{AUC (\%)} & \textbf{F1 (\%)} \\
        \cmidrule(lr){2-3} \cmidrule(lr){4-6}
                       & \multicolumn{1}{c}{\textbf{Image}} & \multicolumn{1}{c}{\textbf{Text}} & \multicolumn{1}{c}{\textbf{Image}} & \multicolumn{1}{c}{\textbf{Text}} & \multicolumn{1}{c}{\textbf{Tag}} &                   &                   &                   \\
        \midrule
        Gemma-3 (27B) \cite{gemmateam2025gemma3technicalreport}   & \(\checkmark\) & \(\checkmark\) & \(\checkmark\) & X          & X          & 63.559 & 62.479 & 53.090 \\
        Gemma-3 (27B) \cite{gemmateam2025gemma3technicalreport}   & \(\checkmark\) & \(\checkmark\) & X          & \(\checkmark\) & X          & 63.559 & 64.337 & 56.270 \\
        Gemma-3 (27B) \cite{gemmateam2025gemma3technicalreport}   & \(\checkmark\) & \(\checkmark\) & \(\checkmark\) & \(\checkmark\) & X          & 61.010 & 60.708 & 51.748 \\
        Gemma-3 (27B) \cite{gemmateam2025gemma3technicalreport}   & \(\checkmark\) & \(\checkmark\) & \(\checkmark\) & \(\checkmark\) & \(\checkmark\) & \textcolor{blue}{\textbf{64.690}} & \textcolor{blue}{\textbf{65.390}} & \textcolor{blue}{\textbf{57.340}} \\
        \midrule
        Mistral Small 3.1 24B & \(\checkmark\) & \(\checkmark\) & X          & \(\checkmark\) & X          & 54.520 & 56.820 & 49.840 \\
        Mistral Small 3.1 24B & \(\checkmark\) & \(\checkmark\) & \(\checkmark\) & X          & X          & 56.500 & 57.420 & 49.340 \\
        Mistral Small 3.1 24B & \(\checkmark\) & \(\checkmark\) & \(\checkmark\) & \(\checkmark\) & X          & 52.540 & 57.160 & 51.720 \\
        Mistral Small 3.1 24B & \(\checkmark\) & \(\checkmark\) & \(\checkmark\) & \(\checkmark\) & \(\checkmark\) & \textcolor{blue}{\textbf{58.470}} & \textcolor{blue}{\textbf{60.800}} & \textcolor{blue}{\textbf{53.630}} \\
        \midrule
        Claude 3 Haiku & \(\checkmark\) & \(\checkmark\) & X          & \(\checkmark\) & X          & 67.230 & 65.490 & 56.060 \\
        Claude 3 Haiku & \(\checkmark\) & \(\checkmark\) & \(\checkmark\) & X          & X          & 68.930 & 64.750 & 53.390 \\
        Claude 3 Haiku & \(\checkmark\) & \(\checkmark\) & \(\checkmark\) & \(\checkmark\) & X          & 70.060 & 68.590 & 59.850 \\
        Claude 3 Haiku & \(\checkmark\) & \(\checkmark\) & \(\checkmark\) & \(\checkmark\) & \(\checkmark\) & \textcolor{blue}{\textbf{72.320}} & \textcolor{blue}{\textbf{72.750}} & \textcolor{blue}{\textbf{65.250}} \\
        \midrule
        GPT-4o      & \(\checkmark\) & \(\checkmark\) & X          & \(\checkmark\) & X          & 72.320 & 69.780 & 60.800 \\
        GPT-4o      & \(\checkmark\) & \(\checkmark\) & \(\checkmark\) & X          & X          & 70.660 & 72.540 & 65.320 \\
        GPT-4o      & \(\checkmark\) & \(\checkmark\) & \(\checkmark\) & \(\checkmark\) & X          & 72.620 & 72.540 & 64.420 \\
        GPT-4o      & \(\checkmark\) & \(\checkmark\) & \(\checkmark\) & \(\checkmark\) & \(\checkmark\) & \textcolor{blue}{\textbf{72.780}} & \textcolor{blue}{\textbf{73.810}} & \textcolor{blue}{\textbf{66.440}} \\
        \midrule
        O3-Mini & X          & \(\checkmark\) & NA         & \(\checkmark\) & X          & 69.770 & 63.359 & 49.289 \\
        O3-Mini & X          & \(\checkmark\) & NA         & \(\checkmark\) & \(\checkmark\) & \textcolor{blue}{\textbf{70.340}} & \textcolor{blue}{\textbf{65.650}} & \textcolor{blue}{\textbf{54.150}} \\
        \midrule
        \textbf{MemeTAG} & \(\checkmark\) & \(\checkmark\) & \(\checkmark\) & \(\checkmark\) & \(\checkmark\) & \textbf{85.540} & \textbf{92.230} & \textbf{84.810} \\
        \bottomrule
    \end{tabular}
\end{table*}
\begin{table}[ht]
    \centering
\caption{Impact of the auxiliary tag-embedding reconstruction\\loss on MemeTAG's performance on the HarMeme dataset}

\label{tab:auxiliary-loss-ablation}
    \begin{tabular}{lcc}
        \toprule
        \textbf{Metric} & \textbf{Without Aux Loss} & \textbf{With Aux Loss} \\
        \midrule
        Accuracy (\%)  & 84.419 & 85.540 \\
        AUROC   (\%)   & 91.910 & 92.230 \\
        F1 Score (\%)  & 83.575 & 84.810 \\
        \bottomrule
    \end{tabular}
\end{table}

\section{Ablation Studies}
In this section, we present ablation studies that address common questions about MemeTAG. We first benchmark our approach against several LLMs on harmful-meme classification. We then ablate the contributions of ATIN and the auxiliary reconstruction loss to overall performance. Additional ablations- including qualitative and quantitative analyses of keyword quality and its impact on MemeTAG, are provided in the supplementary material.
\subsection{Benchmarking Against State-of-the-Art LLMs for Harmful Meme Classification}
To rigorously evaluate the performance of our proposed MemeTAG, we conducted a comparative analysis on the test split of HarMeme dataset  against a selection of prominent large language models considered state-of-the-art as of April 2025. The benchmark models included Gemma-3, Mistral Small 3.1, Claude 3 Haiku, GPT-4o and O3-Mini. To ensure a standardized and fair comparison focused on the specific task of harmful meme detection, and to promote consistent task understanding across models, a carefully constructed prompt was employed uniformly for all models evaluated. This prompt directed the models to analyze both the visual and textual components and perform a binary classification based on a specified definition of harm:\\
\textit{"Analyze the provided meme data. Determine if it qualifies as harmful according to the definition-(causing harm, potentially related to COVID-19, targeting individuals, groups, or society). Output only '1' if harmful, '0' if not harmful."}

Fine-tuning large-scale LLMs on vision--language tasks incurs substantial computational and financial overheads and carries a high risk of overfitting when only small, task-specific datasets are available; accordingly, all LLMs in our study were evaluated in a zero-shot setting. Based on the results presented in Table \ref{tab:llm-comparison}, the MemeTAG model clearly outperformed all other evaluated models by effectively leveraging both image and text modalities. In our evaluation, a clear performance hierarchy emerged among the general-purpose models, with Mistral Small and Gemma-3 consistently yielding lower results across all modality configurations. Mistral Small, in particular, registered the lowest F1-scores, peaking at just 53.630\% even with full multi-modal input. In contrast, GPT-4o and Claude 3 Haiku established themselves as the most capable models in this group, with their performance scaling directly with input complexity. Both models achieved their highest efficacy only when combining Image, Text, and our novel 'Tag' modality, with GPT-4o reaching a top F1-score of 66.440\%. Interestingly, the O3-Mini model, using only text and our tags, was notably competitive, surpassing the fully multi-modal performance of Mistral Small. These findings demonstrate that while the introduction of our 'Tag' modality significantly boosts the performance of not only MemeTAG, but in general to any other model this novelity is incorporated.
\subsection{Impact of ATIN (Aggregate Tag Inference Network) and Auxiliary Reconstruction Loss}
\label{sec:atin-ablation}
To quantify the benefit of our auxiliary reconstruction objective, we perform an ablation study on the HarMeme dataset. Our Aggregate Tag Inference Network (ATIN) produces a concise tag embedding that captures key semantic cues from the input meme. We compare two variants of MemeTAG---one trained with only the classification loss, and one trained with both the classification and auxiliary reconstruction losses---while keeping the rest of the architecture unchanged. As reported in Table~\ref{tab:auxiliary-loss-ablation}, incorporating the auxiliary reconstruction loss (which encourages the model to reconstruct the ATIN embedding) yields an absolute improvement of ~ 1.1 \% in accuracy. This demonstrates that explicitly reconstructing the ATIN-generated embedding strengthens the semantic alignment of the learned features and leads to better overall classification performance across the measured metrics.
\section{Limitations}
While MemeTAG shows promise, it has a few limitations that will greatly benefit from further study. First, because it relies on vision--language backbones such as CLIP trained on web-scale image--text pairs, it inherits their biases: skewed demographic and cultural distributions can teach shortcut correlations where background symbols, clothing, dialect, or identity terms act as proxies for ``harmful'' even when intent is benign. In practice this can raise false positives for specific groups, reclaimed terms, and non-standard spellings, and degrade performance for non-English or low-resource varieties; Like other neural models, it is vulnerable to adversarial tweaks---small edits to images or overlaid text can mislead predictions. And the rapid churn of meme formats and references demands frequent updates and retraining to sustain accuracy without heavy re-engineering.

\section{Conclusion}
 In this paper, we introduced MemeTAG, a dual-objective multimodal framework designed to integrate visual and textual modalities for effective Internet meme classification. Leveraging a robust keyword extraction process and an attention-based Aggregated Tag Inference Network, MemeTAG accurately captures the nuanced semantic interplay present in meme content. Our innovative three-stage training strategy drastically reduces computational overhead, facilitating easy deployment without compromising performance. Experiments on the PrideMM, HarMeme and HMC datasets confirm that MemeTAG outperforms current state-of-the-art methods, achieving significant improvements in accuracy, AUROC, and F1 score. Extensive ablation studies reveal that the performance remains consistent across different keyword-generation models. This robustness is attributed to the ATIN module's ability to suppress irrelevant keywords and focus on those pertinent to the classification task. Future work may explore dynamic keyword generation and further context-aware features to enhance meme classification, as well as real-time deployment for monitoring and mitigating harmful online content.

{\small
\bibliographystyle{memetag_fullname}
\bibliography{references}
}

\end{document}